\documentclass[11pt,a4paper]{article}
\usepackage{emnlp2020}
\usepackage{times}
\usepackage{latexsym,multirow}
\usepackage{graphicx}
\usepackage{courier}
\usepackage{xcolor}
\usepackage[warn]{textcomp}

\usepackage{microtype}

\aclfinalcopy 
\def\aclpaperid{2127}

\title{PharmMT: A Neural Machine Translation Approach to Simplify Prescription Directions}
\author{Jiazhao Li,${}^1$ Corey Lester,${}^2$ Xinyan Zhao,${}^1$ Yuting Ding,${}^2$ \\ 
{\bf Yun Jiang,${}^3$ V.G.Vinod Vydiswaran${}^{4,1}$} \\
${}^1$School of Information; ${}^2$Department of Clinical Pharmacy, College of Pharmacy; \\
${}^3$Department of Systems, Populations, and Leadership, School of Nursing; \\
${}^4$Department of Learning Health Sciences, Medical School \\
University of Michigan, Ann Arbor, MI \\
{\tt \{jiazhaol,lesterca,zhaoxy,dingyt,jiangyu,vgvinodv\}@umich.edu}}

\date{}

\begin{document}
\maketitle
\begin{abstract}
The language used by physicians and health professionals in prescription directions includes medical jargon and implicit directives and causes much confusion among patients. Human intervention to simplify the language at the pharmacies may introduce additional errors that can lead to potentially severe health outcomes. We propose a novel machine translation-based approach, PharmMT, to automatically and reliably simplify prescription directions into patient-friendly language, thereby significantly reducing pharmacist workload. We evaluate the proposed approach over a dataset consisting of over 530K prescriptions obtained from a large mail-order pharmacy. The end-to-end system achieves a \textbf{BLEU score of 60.27} against the reference directions generated by pharmacists, a 39.6\% relative improvement over the rule-based normalization. Pharmacists judged \textbf{94.3\%} of the simplified directions as usable as-is or with minimal changes. This work demonstrates the feasibility of a machine translation-based tool for simplifying prescription directions in real-life.
\end{abstract}

\section{Introduction}

Adverse drug events stemming from medication errors are a vital cause of concern in patient care and are estimated to cost US\$42 billion annually or roughly 1\% of total global expenditure. In the US alone, medication errors cause one death every day and are responsible for over 700,000 visits to the emergency department and over 100,000 hospitalizations each year \cite{Budnitz06, Budnitz11, WHO17a}.

One of the frequent sources of medication errors in the US is the directions on the 1.91 billion electronic prescriptions (e-prescriptions) transmitted annually~\cite{moniz2011addition,odukoya2014prescribing,odukoya2015hidden}. The style and language used in e-prescriptions are highly variable and often filled with medical jargon. For example, in a recent study~\cite{Yang18}, the authors noted that the direction ``\textit{Take 1 tablet by mouth once daily}'' was represented in 832 different ways. The study also found that 10.1\% of e-prescriptions contained incorrect or confusing language. Pharmacists play a vital role as intermediaries between physicians and patients by translating the rich medical jargon in the e-prescriptions written by physicians to patient-comprehensible directions on the prescription labels printed on pill bottles. However, human translation is time-consuming and subject to errors, potentially leading to medication errors and other patient safety risks due to prescription ambiguity.

In this paper, we propose a machine translation-based system, called PharmMT, to simplify the e-prescription directions authored by physicians into patient-friendly language. The system aims to automate the translation and normalization of e-prescription directions and reduce the pharmacists' overall workload. We investigate multiple neural network-based models, including transformer-based models and bi-directional LSTM models, rule-based approaches, and a hybrid model combining neural network-based models with a rule-based backoff. We train and evaluate the proposed system over a dataset of over 530K paired e-prescriptions and their human-translated text, obtained from a large mail-order pharmacy. Using automated measures to evaluate machine translation output, we compare the PharmMT system against a rule-based approach developed based on domain knowledge from pharmacists. Our results show that PharmMT performs significantly better than the rule-based baseline. Manual evaluation by pharmacists also shows a high potential to directly apply the proposed approach in pharmacies to translate e-prescription directions. \\

The contributions of this work are: \\
\indent 1.~We develop a neural machine translation model for simplifying e-prescriptions and build an \textbf{end-to-end system} to generate normalized, patient-friendly, and usable translations.
The model achieves a BLEU score of 60.27 against the reference directions generated by pharmacists, and 94.3\% of simplified prescriptions are judged as usable as-is or with minimal changes by pharmacists. \\
\indent 2.~To the best of our knowledge, our work is the \textbf{first systematic effort} to study neural network models to simplify e-prescription directions. We also developed a rule-based approach as the baseline of this task. \\
\indent 3.~Our work adds \textbf{additional insights} into the limitations of purely automated evaluation metrics of machine translation for domain-adaptive tasks, and offers alternative modes of evaluation.

\begin{table*}[!ht]
\centering
\resizebox{\textwidth}{!}{
\begin{tabular}{|l|l|}
\hline
\multicolumn{1}{|c|}{\textbf{E-prescription direction}} & \multicolumn{1}{c|}{\textbf{Simplified direction}}\\ \hline
2 puffs orally q 4 hrs x90 dys wheeze & Inhale 2 puffs by mouth every 4 hours for 90 days for wheeze \\ 
1 g vaginal mon/tu/th/fr & Insert 1 gram vaginally monday, tuesday, thursday and friday \\ \hline
\multirow{2}{*}{as needed prn; 1 po qd prn} & Take 1 tablet by mouth \textbf{once a day} as needed \\
\multirow{2}{*}{oral one tab po qd prn} & Take 1 tablet by mouth \textbf{every day} as needed \\
 & Take 1 tablet by mouth \textbf{daily} as needed \\ \hline \hline

\multicolumn{1}{|c|}{\textbf{Drug name and strength}} & \multicolumn{1}{c|}{\textbf{Normalized drug name and strength}}\\ \hline
albuterol 90 mcg/inh inhalation aerosol  &   PROAIR HFA AER \\
0.1 mg/g vaginal cream &  ESTRADIOL CRE 0.01\% \\
traMADol 50 mg tablet & TRAMADOL HCL TAB 50MG \\ \hline

\end{tabular}
}
\caption{\textbf{Top}: Example pairs of e-prescription directions and corresponding simplified text. Variations of these directions exist on both sides. \textbf{Bottom}: Drug name and strength information in the original and normalized forms.}
\label{source_target_exmaple}
\end{table*}

\section{Related work}

Prior work on automated approaches for translating e-prescription directions mainly focused on information extraction models relied on handwritten rules or linguistic signals found in prescription free text. 
Tools such as MetaMap~\cite{aronson2010overview} and MedLEE~\cite{friedman1996web} extract and organize clinical information in text documents using external knowledge sources, such as the Unified Medical Language System (UMLS). Other systems, such as FABLE~\cite{tao2018fable}, employed a conditional random fields-based model to recognize medication entities. Other researchers have proposed rule-based approaches to normalize and simplify directions using task-specific knowledge such as common abbreviations used in prescriptions~\cite{qenam2017text,kandula2010semantic}.

While machine translation-based approaches have not yet been proposed for translating e-prescription directions, prior works such as \cite{yolchuyeva2018text,shardlow-nawaz-2019-neural,van2019evaluating} have suggested solving machine translation tasks without the need for explicitly-defined rules. Neural machine translation (NMT) models have been shown to be able to learn contextual rules automatically from large corpora and produce higher quality translations~\cite{bahdanau2014neural,wu2016google,lee2017fully}. Other researchers, such as~\cite{aw2006phrase,xu2016optimizing}, have shown that while statistical machine translation methods mainly focused on lexical rules to minimize sentence complexity, NMT models could capture richer syntactic information~\cite{shi-etal-2016-string}.

Researchers studying deep neural network models have explored multiple encoder-decoder frameworks, such as Transformer-based networks~\cite{NIPS2017_7181}, and Recurrent Neural Networks (RNN), including Long Short-Term Memory (LSTM) models~\cite{hochreiter1997long} and Gated Recurrent Units (GRU)~\cite{chung2014empirical}. RNN units have been used to encode source sentences into fixed-length representations and then decoded into reference sentences~\cite{DBLP:journals/corr/ChoMGBSB14}. Models with deep LSTM-based RNN units have shown the benefits of using deeper structure~\cite{DBLP:journals/corr/WuSCLNMKCGMKSJL16}. In other works, \cite{DBLP:journals/corr/GehringAGYD17} introduced a convolution neural network with an attention-based mechanism to learn long-range dependency. In recent work, \cite{NIPS2017_7181} developed a novel Transformer-based architecture using attention mechanism without recurrence or convolution, resulting in a state-of-the-art performance for many natural language processing tasks, such as recognizing textual entailment, sentiment analysis, and natural language inference.~\cite{raffel2019exploring,lan2019albert,devlin2018bert}

\section{Simplifying e-prescription directions} \label{Simplify}
We frame the challenge of simplifying e-prescription directions as a machine translation task from physician-authored directions (``\textbf{source}'') to patient-facing text authored by pharmacists (``\textbf{reference}''). This monolingual translation task focuses on replacing highly-abbreviated medical jargon with patient-friendly vocabulary, simplifying cryptic expressions, and normalizing them so that they can be used with minimal changes by the pharmacists. Table~\ref{source_target_exmaple} (top panel) shows three examples of e-prescriptions and their corresponding simplified directions. 

E-prescription directions consists of specific components related to the prescribed drug, viz.,~\textit{dosage}, \textit{form}, \textit{route}, \textit{duration}, \textit{frequency}, and \textit{reason} for prescribed use. For example, the e-prescription ``\textit{2 puffs orally q 4 hrs x90 dys wheeze}'' specifies that the patient should inhale 2 (\textit{dosage}) puffs (\textit{form}) by mouth (\textit{route}) every 4 hours (\textit{frequency}) for 90 days (\textit{duration}) for wheezing (\textit{reason)}. While not all components are present in every e-prescription direction, some components are critical and need to be stated explicitly. The \textit{name} and \textit{strength} of the prescribed drug are also available as auxiliary information. Examples of drug names and strengths are shown in the bottom panel of Table~\ref{source_target_exmaple}.

\begin{figure}[t]
\center{\includegraphics[width=0.9\columnwidth]{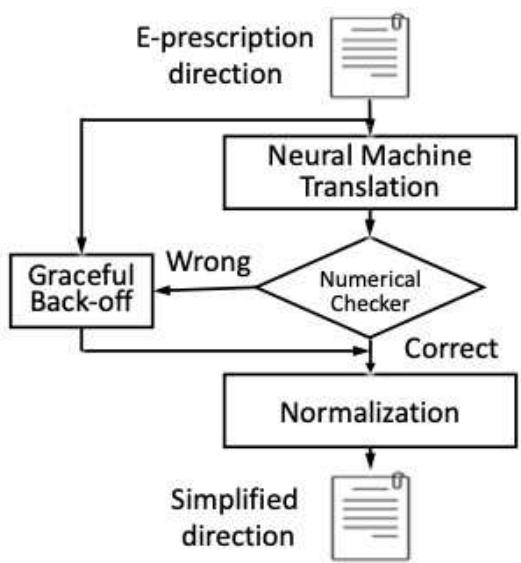}}
\caption{Schematic diagram of the PharmMT system}
\label{Model}
\end{figure}

While we formulate the challenge as a machine translation task in the pharmacy domain, key desiderata for the automated approach are to preserve the accuracy and consistency of the critical components in a prescription. To achieve this, we develop an end-to-end system called \textbf{PharmMT}, consisting of three stages: neural machine translation, numerical check and graceful backoff, and normalization, as depicted in Figure~\ref{Model}.

\subsection{Neural Machine Translation (NMT)}\label{NMT}
The primary component of the proposed approach is a sequential model that ``translates'' physician-authored e-prescription text to normalized, patient-friendly language using an NMT framework. NMT models map a source sequence, $\mathbf{x:}~x_1, x_2, \ldots, x_n$ into a reference sequence, $\mathbf{y:}~y_1, y_2, \ldots, y_m$ by maximizing the conditional probability $p(\mathbf{y}|\mathbf{x})$ using an Encoder-Decoder framework~\cite{DBLP:journals/corr/Neubig17}. One such model is a recurrent sequence-to-sequence model, which consists of a bidirectional LSTM model~\cite{schuster1997bidirectional} with global attention as the encoder and a forward-sequence LSTM model~\cite{hochreiter1997long} as the decoder. Both encoder and decoder stages are configured as multi-layer models, with a hidden state in each layer, to sufficiently capture the deep semantic components in the e-prescription \cite{barone2017deep}.

To compare against the performance of the recurrent sequence-to-sequence model, we also trained an attention-based transformer model~\cite{NIPS2017_7181}. Position embedding was enabled to capture sequence information and provide similar architectural complexity as a recurrent network. Both models were developed using the OpenNMT framework~\cite{klein-etal-2017-opennmt, klein-etal-2018-opennmt}, and the dropout probability was adjusted to prevent over-fitting~\cite{srivastava2014dropout}. 
Additional experimental details can be found in Section~\ref{Training_process}.

\subsubsection{Augmenting auxiliary information}\label{Auxiliary}

As described in Section~\ref{Simplify}, e-prescriptions contain auxiliary information on the drug name and strength. While the primary task is to simplify just the direction, access to the auxiliary information may help distinguish directions based on the context associated with drugs. We hypothesize that the auxiliary information will improve the neural machine translation models to simplify directions. This hypothesis also matches with real-life information available to pharmacists. To test this hypothesis, we prepend the drug name and strength information to the ``source'' direction before training the models. We evaluate the updated model on the original task of simplifying just the directions. 
\subsubsection{Pre-trained word embeddings} \label{sec:wordEmbed}
The input representation is a pre-trained word embedding layer to allow for similar representations of words in similar contexts. Word embeddings can capture fine-grained semantic and syntactic word relationships and, in turn, allow for a better initialization for gradient optimization. We explored static pre-trained embedding models and compared them against a randomly-initialized representation vector. 
The first one was the general-domain GloVe word embeddings, pre-trained on the Wikipedia and Gigaword corpora~\cite{pennington2014glove}. The second one was clinical domain-adaptive word embeddings that we trained on MIMIC-III, a large corpus of clinical notes~\cite{johnson2016mimic} and a dataset of pharmacy directions~\cite{Pharmacy_2020}.
We hypothesize that domain-adaptive word embeddings would outperform both the general-domain embeddings and randomly-initialized vector embeddings.

\subsubsection{Learning ensemble models}\label{Ensemble}

The primary motivation of ensemble learning over neural network models is to improve the final model's robustness to the variations introduced in parameterized modules because of dropout probability and random seeding. In an ensemble model, the final distribution of the output dictionary is computed by averaging the output distributions from the trained models in the ensemble during the inference phase.

\subsection{Numerical check}\label{sec:NER_checker}

Once the machine translation module generates the simplified candidate directions, the candidates are checked for consistency of key components of the prescription. Numerical components -- including \textit{dosage}, \textit{frequency}, and \textit{duration} -- 
are critical in prescriptions. Medication under-dosage often leads to poorer health outcomes, while over-dosage can be severe, even fatal. 

The correctness of the numerical components in the simplified directions is checked by comparing against the source e-prescription. 
We incorporated two different numerical checking strategies -- \textbf{Token-based} and \textbf{NER-based}. In the token-based checking, all numeric tokens that appear in the simplified direction were checked against numeric tokens in the source direction. The bag of tokens approach helps tag any simplified direction that ``makes up'' numeric values in a key component.

On the other hand, the token-based checker can also generate faulty consistency claims, for example, when the simplified direction swaps a \textit{dosage} term with the \textit{frequency}. To overcome this, we incorporated a pre-trained medication NER model~\cite{XinyanNER} to tag \textit{dosage}, \textit{frequency}, and \textit{duration} components in both source and simplified directions, and compared them component-by-component. The NER model was trained over a medication extraction task~\cite{henry20202018} and achieved an overall F1 score of 0.9571 over all medication components.

\begin{table}[!t]
\begin{center}

\begin{tabular}{|l|l|l|} \hline
\multicolumn{1}{|c|}{\textbf{Fields}} & \multicolumn{1}{c|}{\textbf{Original}}  & \multicolumn{1}{c|}{\textbf{Normalized}} \\ \hline
\multirow{3}*{Action}  & \multirow{3}*{(missing)} &  take \\
             & & inject \\
             & & inhale \ldots\\ \hline
\multirow{3}*{Dosage} & one and half & 1.5 \\ 
                 & 1 1/2 & 1.5 \\ 
                 & one (1) & 1 \\ \hline
\multirow{3}*{Form} & tab, tabs & tablet \\ 
             & cap & capsule \\
             & in, inj, injctor & injection \\ \hline
\multirow{2}*{Route}  & orally, by oral & by mouth   \\ 
        & sq, subcutaneous & under the skin   \\ \hline
\multirow{2}*{Frequency} & {qd}  & {every day}  \\                                   
        & {bid}  & {twice a day}  \\ \hline
Duration &  x3 week & for 3 weeks  \\ \hline
\end{tabular}
\caption{Sample normalization and simplification rules}
\label{TableNormalization}

\end{center}
\end{table}

\subsection{Graceful backoff}

If the simplified direction is deemed consistent after the numeric check, the direction is considered as the final candidate for normalization. However, if the numeric check fails, the NMT output is discarded and the original source direction is used as the final candidate. This graceful backoff represents a trade-off between information accuracy and good language model performance. 

\subsection{Normalization}\label{Normalization}

Before the candidate direction from the numerical check phase is finalized, the candidate text undergoes pharmacy-specific post-processing and simplification. 
Two pharmacists identified common linguistic patterns in pharmacy directions, which were coded into normalization rules. Highly-abbreviated medical jargon was replaced, for example, by replacing the Latin term \textit{bid} with its synonymous phrase \textit{twice a day}. Action verbs appropriate for the form of the drug, such as \textit{inject} (syringe), \textit{inhale} (nebulizer), and \textit{take} (capsule), were added. Numerical values in words or fractions were converted to digits (e.g.~\textit{1 1/2} was converted to \textit{1.5}). Abbreviations and other common medical expressions were normalized into standard and patient-friendly variants, e.g.~\textit{inj} or \textit{injector} to \textit{injection}. Overall, the normalization step included more than 300 rules. Table~\ref{TableNormalization} shows examples of these normalization and simplification rules. The normalization module also served as the rule-based baseline.

\begin{table}[!t]
\begin{center}
\resizebox{\columnwidth}{!}{
\begin{tabular}{ |c|c|c|c| } 
\hline
\textbf{Data set}  & \textbf{Train}     & \textbf{Validation}     &  \textbf{Test}         \\ \hline
Original        & 318,594   & 79,648  &  132,747         \\ \hline
Deduplicated  & 318,594   & 15,625  &  36,652          \\ \hline
\end{tabular}
}
\caption{Data set sizes after removing duplicates.}
\label{dataset_distribution}
\end{center}
\end{table}

\section{Experimental setup}
\subsection{Data set description}
The e-prescription corpus used in this study consists of all e-prescriptions dispensed by an online outpatient mail-order pharmacy from their dispensing software from January 2017 to October 2018. The corpus consists of 530,988 e-prescriptions received from 65,139 unique physicians from all fifty US~states~\cite{Pharmacy_2020}.

Each e-prescription direction in the data set is paired with the corresponding simplified text authored by a mail-order pharmacy team member. Table~\ref{source_target_exmaple} shows some example e-prescription directions and the corresponding simplified text. 

In addition to the e-prescription, each direction in the corpus also contains auxiliary information about the name and strength of the drug. However, similar to the directions, physician-authored information often included chemical or ingredient names for the drug, while the pharmacist-translated direction contained generic drug names or brand names.

In all, there are 120,402 unique e-prescription directions and 83,823 unique pharmacist-authored directions in the dataset. The difference in these numbers is due to the diverse writing styles of physicians and pharmacists. On an average, there were 6.33 e-prescription directions mapped to a single pharmacist-authored direction; while one e-prescription direction mapped, on an average, to 4.41 different pharmacist-authored directions.

We split our data into train, validation, and test sets. To avoid information leak during the evaluation, we remove all duplicates appearing in more than one set, but retain those appearing within a single set. Table~\ref{dataset_distribution} summarizes the distribution of the instances over the three sets. None of the instances in the validation or test sets were used during the training phase.

\begin{table}[!t]
\begin{center}
\resizebox{\columnwidth}{!}{
\begin{tabular}{ |c|c|c|c| } 
\hline
\multirow{2}{*}{\textbf{Embeddings}} & \textbf{Vocab.}  &  \textbf{Source} & \textbf{Reference}     \\
 & \textbf{size} & (n = 11,643) & (n = 7,358) \\ \hline
Clinical & 498,677 & 10\% & 6.69\%      \\
General & 400,000 & 73.75\% & 60.29\%      \\\hline
\end{tabular}
}
\caption{Out-of-vocabulary (oov) ratio for pre-trained word embedding models. General-domain word embeddings show a large oov gap against our dataset.}
\label{Vocabulary}
\end{center}
\end{table}

\subsection{Training process}\label{Training_process}
To prepare the data for training, the source e-prescription directions and reference pharmacist-authored directions are prepended with the drug name and strength, as described in Section~\ref{Auxiliary}.

\begin{table*}[!t]
\centering
\begin{tabular}{|l|c|c|}
\hline
\multicolumn{1}{|c|}{\textbf{E-prescription direction}} & \multicolumn{1}{c|}{\textbf{{\fontsize{8}{10}\selectfont BLEU}}} & 
\multicolumn{1}{c|}{\textbf{{\fontsize{8}{10}\selectfont METEOR}}}\\ \hline

location: both eyes. 1 drop into each eye at bedtime. \textit{[Source]} && \\
apply 1 drop into each eye at bedtime. \textit{[NMT]} &0&0.77\\
instill 1 drop into both eyes at bedtime.  \textit{[Reference]} &  & \\\hline
1 puff once daily per dr.~jones. \textit{[Source]} & & \\
inhale the contents of one capsule by mouth once daily using handihaler. \textit{[NMT]} & 0.38 & 0.90 \\

use 1 inhalation by mouth once daily. \textit{[Reference]} & & \\\hline

\multicolumn{1}{|c}{\textbf{Evaluation Limitation}} & \multicolumn{2}{c|}{} \\ \hline

take \textbf{1} tablet by mouth every morning \textbf{and} every evening. \textit{[Reference]} &&\\
take 1 tablet by mouth every morning \textbf{\&} every evening. \textit{[NMT$_{1}$]} & 0.70 & 0.91 \\
take \textbf{10} tablets by mouth every morning and every evening. \textit{[NMT$_{2}$]} & 0.70 & 0.98 \\
\hline
\end{tabular}
\caption{\textbf{Top}: Comparison of BLEU and METEOR scores on two examples. METEOR scores show higher correlation with human judgement due to flexible matching. \textbf{Bottom}: Both variations of NMT model outputs have similar BLEU and METEOR scores. However, The first replaced an `\textit{\&}' with `\textit{and}', while the second had a critical dosage error (`\textit{10}', instead of `\textit{1}').}
\label{Metric_camparison}
\end{table*}

\paragraph{Pre-trained word embeddings:} We tested two static pre-trained word embeddings -- general-domain GloVe embeddings and a second one trained explicitly over two clinical domain corpora -- against a randomly-initialized representation. The out-of-vocabulary rate is shown in Table~\ref{Vocabulary}. While only 10\% of the source words and 6.69\% of the target words had no word embeddings in the clinical-domain word embeddings, the out-of-vocabulary ratio was 7 to 9 times higher in the general-domain word embeddings.

\paragraph{Model configuration:} The number of layers in the encoder and decoder stages of the Bi-LSTM, LSTM-based, and Transformer-based models was empirically chosen from the set \{2, 4, 6, 8\}. The length of hidden states was chosen from the set \{128, 256, 512\}. In the following description, we denote the number of layers (i.e., Transformer blocks) as L, the hidden size as H, and the number of self-attention heads as A, consistent with the BERT notation~\cite{devlin2018bert}. After choosing the hyper-parameter based on highest BLEU score, we primarily report results on two best performance architectures: \textbf{Bi-LSTM/LSTM-based} model (L = 4, H = 256, Dropout = 0.4) and \textbf{Transformed-based} model (L = 4, H = 128, A = 2, Dropout = 0.2). We trained our settled models on a single Tesla V100. It took 4.67 hours to train the LSTM-based models with 12.66 million parameters, and 3.33 hours for the Transformer-based model with 9.27 million parameters.

\subsection{Evaluation metrics}
\paragraph{Automatic evaluation:}
We evaluated the translation model using two automated candidate-reference comparison metrics: BLEU~\cite{papineni2002bleu} and METEOR ~\cite{denkowski:lavie:meteor-wmt:2014}. The BLEU-4 score is the most popular metric used to evaluate the similarity between the candidate text and human reference in machine translation tasks~\cite{sutskever2014sequence, koehn2003statistical, lipton2015critical}. It is computed as the geometric mean of precision of unigram, bigram, trigram, and 4-gram matches between the candidate and reference texts. In contrast to the ngram-based overlap in BLEU-4, the METEOR score considers unigram alignment between candidate and reference texts. This results in more flexible matching, including stem match and synonym match using WordNet~\cite{miller1995wordnet}. METEOR scores are known to achieve a better correlation with human judgment ~\cite{banerjee2005meteor}. 

Sample comparisons of two metrics in Table~\ref{Metric_camparison} (top panel) show the limitation of using the BLEU score to compare pharmacy instructions, while also showing the feasibility of using METEOR score as a viable alternative. The bottom panel of Table~\ref{Metric_camparison} highlights the limitation of both metrics in checking the consistency of prescription components, and is discussed in more detail in Section~\ref{MetricLimitation}.

\begin{table}[!t]
\centering
\resizebox{\columnwidth}{!}{
\begin{tabular}{|l|c|c|}\hline
\multicolumn{1}{|c|}{\textbf{NMT Models}} & \textbf{BLEU} & \textbf{METEOR} \\\hline

TransF-TransF : Random   &  59.09\footnotesize{$\pm$0.08} & 80.10\footnotesize{$\pm$0.06} \\
TransF-TransF : GloVe &  62.36\footnotesize{$\pm$0.10} & 80.45\footnotesize{$\pm$0.05} \\
TransF-TransF : Clinical    & 63.23\footnotesize{$\pm$0.10}  & 80.89\footnotesize{$\pm$0.17} \\ 
\multicolumn{1}{|r|}{+ Ensemble} & 64.19\footnotesize{$\pm$0.31} & 81.05\footnotesize{$\pm$0.12} \\ \hline
BiLSTM-LSTM : Random    &  64.63\footnotesize{$\pm$0.06} & 81.14\footnotesize{$\pm$0.30} \\ 
BiLSTM-LSTM : GloVe    & 65.78\footnotesize{$\pm$0.07}  & 81.62\footnotesize{$\pm$0.11} \\ 
BiLSTM-LSTM : Clinical   &  66.02\footnotesize{$\pm$0.03} & 82.32\footnotesize{$\pm$0.20} \\ 
\multicolumn{1}{|r|}{+ Ensemble} &  \textbf{66.61\footnotesize{$\pm$0.29}} & \textbf{82.73\footnotesize{$\pm$0.10}} \\  \hline
Rule-based baseline & 43.19 & 68.59  \\ \hline
\end{tabular}
}
\caption{\label{NMT_Result_table}Comparison of NMT models with different word embeddings approaches, against a rule-based baseline. An ensemble BiLSTM-LSTM model with clinical word embeddings achieves the highest BLEU and METEOR scores.}
\end{table}

\paragraph{Manual evaluation:}
While automated metrics such as BLEU and METEOR are commonly used to evaluate machine translation tasks, they do not sufficiently evaluate the clinical usability of the candidate texts. Hence, in addition to the automated evaluation, we asked two pharmacist trainees to evaluate 300 pairs of e-prescription directions and corresponding simplified directions randomly sampled from the test set. The pharmacist trainees were asked to classify the direction pairs into one of three categories -- \textbf{Correct:} all information in the simplified direction was correct; \textbf{Missing:} the simplified direction was correct, but missed some essential information; and
\textbf{Wrong:} the simplified direction contained fundamental errors that needed to be corrected.
A pharmacist expert resolved labeling disagreements. 
This manual evaluation simulates the human effort undertaken in real-life at the pharmacies to simplify e-prescription directions.

\begin{table*}[!ht]
\centering
\begin{small}

\begin{tabular}{|l|p{1.55cm}|p{4.5cm}|p{4.5cm}|p{3.2cm}|}
\hline
\multicolumn{1}{|c|}{\#} &
\multicolumn{1}{|c|}{\textbf{Issues}} & \multicolumn{1}{|c|}{\textbf{E-prescription}} & \multicolumn{1}{|c|}{\textbf{Rule-based baseline}} & \multicolumn{1}{|c|}{\textbf{PharmMT}} \\ \hline
1 & Contextual ambiguity & 1/2 tab bid orally \textbf{90}. & Take 0.5 tablet by mouth twice a day \textbf{90} & Take 0.5 tablet by mouth twice a day \textbf{for 90 days.} \\ \hline
2 & Word sense ambiguity & \textbf{spray} 1 \textbf{spray(s)} 4 times a day by intranasal route as needed for 90 days.  & Use \textbf{spray} 1 \textbf{spray} 4 times a day in the nose route as needed for 90 days. & Use 1 \textbf{spray} in the nose 4 times a day as needed .\\\hline
3 & Re-ordering components & tablets by mouth daily; \textbf{3.5 tab 7 mg.} & Take tablets by mouth daily; \textbf{3.5 tablet 7 mg}. & Take \textbf{3.5 tablets} by mouth daily. \\\hline
4 & Handling misspelling & one tablet by mouth \textbf{oce} daily.  & Take one tablet by mouth \textbf{oce} daily. & Take 1 tablet by mouth \textbf{once} a day. \\\hline
5 & Informal abbreviations & 1 puff \textbf{aero pow br act} bid.  & Inhale 1 puff \textbf{aero pow br act} twice a day. & Inhale 1 puff \textbf{by mouth} twice a day. \\\hline
\end{tabular}
\end{small}
\caption{Examples comparing PharmMT model against a rule-based baseline, highlighting potential issues with rule-based approaches.}
\label{base-line_exmaple}
\end{table*}

\section{Results}
We present our results first by evaluating the NMT module by comparing the two proposed architectures using the automated BLEU and METEOR scores. After finalizing the best performing NMT model, we compare its results against a rule-based baseline and demonstrate the significance of the individual stages through an ablation study. Finally, we report the performance of the end-to-end system based on the manual evaluation.

\subsection{Evaluating Neural Machine Translation module}\label{Eva_NMT}

We compared two classes of NMT models -- \textbf{Transformer-based} model and \textbf{BiLSTM/LSTM-based} model under different pre-trained word embeddings. The results are summarized in Table~\ref{NMT_Result_table}. The reported values are mean and standard deviation over ten independent iterations of training and validation using the same model hyper-parameters.

Both automated metrics were consistent in their ranking of the systems. On both metrics, LSTM-based models outperformed transformer-based models. One possible explanation for these results is that although transformer-based models are better at capturing long-range dependencies, their advantage is nullified by the relatively short sentences in this task. The average length of e-prescription directions is 10.42$\pm$4.55 tokens. 

Models using the pre-trained clinical-domain word embeddings led to the highest performance on both metrics and were statistically better than models that used general-domain word embeddings. Models using the randomly-initiated word representation performed the worst. The model performance improved further when ensemble learning was applied. The ensemble BiLSTM / LSTM model with clinical domain-adaptive word embeddings achieved the highest overall BLEU score of 66.61$\pm$0.29 and the METEOR score of 82.73$\pm$0.10.

We also note that in Section~\ref{sec:wordEmbed}, we stated our hypothesis that domain-adaptive word embeddings would outperform both the general-domain embeddings and randomly-initialized vector representations. This hypothesis was shown to be valid in our results in Table~\ref{NMT_Result_table}. Instead of the static, but domain-adaptive, word embeddings explored in this work, other alternatives such as contextual word embeddings could also be used, including BioBERT~\cite{BioBERT} and ClinicalBERT~\cite{ClinicalBERT}.

\subsection{Comparison to rule-based baseline}
Next, we compared the translated directions generated by the PharmMT model against a rule-based baseline in which the e-prescriptions are passed directly through the Normalization module to produce the outputs. Table~\ref{base-line_exmaple} shows some examples of rule-based translation against PharmMT output. These examples highlight three potential issues of rule-based systems, viz., handling ambiguity, reordering of direction components, and sensitivity to misspelled tokens and abbreviations. 

In the first example, the rule-based approach fails to recognize `\textit{90}' as \textit{duration} without any contextual clues, whereas PharmMT correctly normalizes it as `\textit{for 90 days}'. Similarly, in the second example, the rule-based approach fails to identify `\textit{spray}' as both an action verb and the \textit{form} token. This highlights the difficulty in manually curating rules to cover all ambiguous cases.

Second, while rule-based approaches are very effective when the order of prescription components is as expected, they do not handle reordered components well. The third example in Table~\ref{base-line_exmaple} shows one such instance, where unlike the rule-based approach, the PharmMT model correctly reorders the direction components by inserting \textit{dosage} tokens (`\textit{3.5 tablets}') before the \textit{form} token, `\textit{tablet}'. 

Finally, rule-based approaches are also more sensitive to misspelled tokens and informal abbreviations, compared to PharmMT, as shown in the final two examples in Table~\ref{base-line_exmaple}. 

\begin{table}[!t]
\centering

\resizebox{\columnwidth}{!}{
\begin{tabular}{|l|c|c|}\hline
\multicolumn{1}{|c|}{\textbf{Model variations}} & BLEU & METEOR \\ \hline
(1):~Rule-based Baseline (Normalizer) & 43.19 & 68.59  \\

(2):~Best NMT model &  67.21 & 82.90  \\
(3):~Best NMT model w/o Auxiliary & 63.59 & 80.83  \\ 
(4):~(2) + Backoff & 66.24  &  79.59 \\
(5):~\textbf{PharmMT} [same as (4) + (1)] &  \textbf{60.27} & \textbf{76.11} \\ 
(6):~(5) - Backoff [same as (2) + (1)] & 60.29 & 76.20  \\
\hline
\end{tabular}
}

\caption{\label{NMT_Result_ablation} Ablation study results showing model performance taking components out one-at-a-time, from the best NMT model to the end-to-end system, PharmMT (in bold).}
\end{table}

\subsection{Results of the ablation study}\label{abla}

The best performing NMT model achieved a BLEU score of 67.21 and a METEOR score of 82.90. After finalizing the best NMT model, the significance of the remaining components is shown using an ablation study. The results are summarized in Table~\ref{NMT_Result_ablation}. 

\paragraph{Removing auxiliary information:} Without augmenting auxiliary drug information, the performance of the NMT model drops by 5.4\% on the BLEU score and 2.5\% on the METEOR score. These results indicate that augmenting directions with auxiliary drug information helps improve the overall performance, as hypothesized in Section~\ref{Auxiliary}.

\paragraph{Graceful backoff and normalization:}
Adding the NER-based numeric check and resorting to graceful backoff, when necessary, reduces the overall scores on the automated metrics (BLEU: 66.24, METEOR: 79.59). Adding normalization decreases both metrics even further (BLEU: 60.27, METEOR: 76.11). This performance drop is because the normalization-based approaches tend to produce reference direction texts influenced heavily by preference rules from pharmacists. So, while the resultant directions are more readable, patient-friendly, and preferred by pharmacists, they do not accurately reflect the preferred style in the original reference corpus. We expand on this further in Section~\ref{Eva_Norm}.

\subsection{Manual evaluation of end-to-end system}\label{human_eval}
The final output of the end-to-end system was evaluated by domain experts. Of the 300 pairs of e-prescription directions and their corresponding simplified texts, 86.7\% (n=260) were labeled as Correct, 7.6\% (n=23) as Missing; and 5.7\% (n=17) as Wrong. The missing errors were primarily related to missing adjectives and adverbs (e.g.,~\textit{transdermal}, \textit{slowly}), or typographic omissions, such as brackets. The incorrect errors were primarily related to special directions (e.g.~taking medications \textit{before meals}, \textit{with food}), formatting issues with dosage (e.g.~\textit{10-12}), or complex directions based on days of the week (e.g.~\textit{every day except Sundays}). These results show that in \textbf{94.3\% instances}, the simplified output can be used as-is or after minimal changes to add the missing elements.  

\section{Discussion}
\subsection{Error analysis}
We further analyzed all non-`Correct' instances (n=17; 5.7\%) identified during the manual evaluation. A major class of errors was complicated instruction and language patterns in the e-prescription. On average, these directions were 16.86 words long, compared to an average of 12.57 words for `Correct' instances. For example, one direction that was not simplified correctly was: \textit{30 units with meals plus ssi 150-200 2 units, 201-250 4 units, 251-300 6 units, 301-350 8 units, greater than 351 10 units.} This direction instructed patients to change dosage depending on the sliding scale for insulin (\textit{ssi}). 

Based on the hypothesis that shorter directions will have simpler language patterns, we evaluated a subset of test instances (n=11,977; 32.6\%) that were under 12 words long. This `shorter length' subset achieved an aggregate BLEU score of 71.14, while the complementary `longer length' subset managed an aggregate BLEU score of 64.02. 

\begin{table}[!t]
\centering
\resizebox{\columnwidth}{!}{
\begin{tabular}{|l|c|c|c|c|}\hline
 & Dosage & Frequency & Duration & Combined \\
 & (n=22,099) & (n=8,256) & (n=2,879) & (n=23,237) \\ \hline
NER & 2,570 & 1,451 & 238 & 4,027 \\ \hline
Token & - & - & - & 1,390 \\ \hline

\end{tabular}
}
\caption{\label{tab:NER_checker} Number of instances marked as inconsistent by the numeric checkers. NER-based checker flagged more inconsistent instances than NER-based checker.}
\end{table}

\subsection{Numeric checker and Graceful backoff}\label{Eva_NER}
We further investigated the performance of the numeric checker. The number of instances marked as inconsistent by the two approaches is summarized in Table~\ref{tab:NER_checker}. The stricter, NER-based numeric checker flagged 4,027 (17.33\%) instances as inconsistent, while the token-based checker flagged only 1,390 (5.98\%) of instances as inconsistent.

\subsection{Need for normalized reference}\label{Eva_Norm}
The normalization process is heavily influenced by the style preferred by pharmacists coding the normalization rules. Since the preferred style of the team that generated the original reference differed from the expert pharmacists in our team, the original reference data were themselves not normalized. This led to a reduction in BLEU scores when normalization was added (see Table~\ref{NMT_Result_ablation}). 

To understand how well our trained models could perform on this corpus, we created a normalized version of the reference corpus. Using this as the gold reference, the original reference corpus has a BLEU score of 82.48, and that of the PharmMT system was 62.68 (see Table~\ref{BLEUNorm}). The table also shows the ratio of test instances that are normalized to indicate how \textit{close} the output is to the normalization rules: the lower the ratio, the closer it is. The results indicate that the NMT model learned more latent rules from the train data set than the hand-crafted normalization rules, while having a higher BLEU score.

\begin{table}[!t]
\centering
\resizebox{\columnwidth}{!}{
\begin{tabular}{|l|c|c|}
\hline
\multicolumn{1}{|l|}{\textbf{Against normalized reference}} & \textbf{BLEU} & \textbf{Ratio} \\ \hline
Rule-based Baseline (Normalizer) & 47.60 & 86.58\%  \\
Best NMT model & 62.68 & -  \\ 
Best NMT model + Normalizer & 71.33 & 30.81\% \\
PharmMT & 71.14 & 36.01\% \\ \hline
Reference (upper bound) & 82.48 & 42.36\%  \\ \hline

\end{tabular}
}

\caption{\label{BLEUNorm} Performance against normalized reference}
\end{table}
\subsection{Limitation of BLEU and METEOR} \label{MetricLimitation}
Automated metrics such as BLEU and METEOR can only evaluate the translation results using a linguistic, token-level approach, but fail to capture the nuanced semantic-level information. However, in prescription directions, different words contain \textbf{unequal} useful information, and hence should be re-weighted during the evaluation process. As we noted in Section~\ref{sec:NER_checker}, consistency of key information is vital for patient safety. For example, in the translation shown in bottom of Table~\ref{Metric_camparison}, both machine translation outputs \textit{NMT$_1$} and \textit{NMT$_2$} have only one token different from the reference. But, \textit{NMT$_1$} is labeled as `Correct' in the manual evaluation while \textit{NMT$_2$} is labeled as `Wrong' because of a serious error on \textit{dosage}. However, both translations got similar BLEU and METEOR scores. In the future, we will focus on improving information consistency while maintaining high model performance. 

\section{Conclusion}
We proposed and developed a machine translation-based approach, called PharmMT, to simplify e-prescription directions. We systematically evaluated the individual stages and the overall approach over a large mail-order pharmacy data corpus. Our results showed that an ensemble model with a bi-directional LSTM encoder and an LSTM decoder, trained over a clinical-domain word embedding representations, achieved the best overall BLEU score of 60.27. NER-based numeric check and graceful backoff ensure information consistency and the normalization stage helps generate patient-friendly directions. Qualitative evaluation by domain experts showed that 94.3\% of the simplified directions could be used as-is or with minimal changes. These results indicate that the proposed approach could be deployed in practice to automate the simplification of prescription directions.

\section*{Acknowledgment}
This work was partially supported by an MCubed grant from the University of Michigan, titled ``TRANSCRIBE: Development of an algorithm to automatically transcribe e-prescription directions in the pharmacy'' and faculty startup grants.

\bibliographystyle{main_acl_EMNLP20}
\bibliography{main_acl_EMNLP20.bib}

\end{document}